\documentclass{article}

\usepackage{PRIMEarxiv}

\usepackage[utf8]{inputenc} 
\usepackage[T1]{fontenc}    
\usepackage{hyperref}       
\usepackage{url}            
\usepackage{booktabs}       
\usepackage{amsfonts}       
\usepackage{nicefrac}       
\usepackage{microtype}      
\usepackage{lipsum}
\usepackage{fancyhdr}       
\usepackage{graphicx}       
\graphicspath{{media/}}     

\usepackage{amsmath}
\usepackage{diagbox}
\usepackage{tabularx}
\usepackage{tikz}
\usepackage{bm}
\usepackage{url}
\usepackage{multirow}

\newcommand{\nsamp}{{N}}
\newcommand{\npar}{{n_\theta}}
\newcommand{\nx}{{n_x}}
\newcommand{\ny}{{n_y}}
\newcommand{\nin}{{n_u}}
\newcommand{\M}{\mathcal{M}}
\newcommand{\F}{\mathcal{F}}
\newcommand{\G}{\mathcal{G}}
\newcommand{\N}{\mathcal{N}} 
\newcommand{\D}{\mathcal{D}}
\newcommand{\R}{{\mathbb{R}}}
\newcommand{\sys}{{\mathcal{S}}}
\newcommand{\tr}{{\rm{tr}}}
\newcommand{\tst}{{\rm{tst}}}
\newcommand{\transf}{{\rm{xf}}}
\newcommand{\eval}{{\rm{ev}}}
\newcommand{\lin}{{\ell}} 
\newcommand{\ops}{\rm{ops}}
\newcommand{\nominal}{{\rm{nl}}}

\newcommand{\norm}[1]{\left\lVert#1\right\rVert}
\newcommand{\tvec}[1]{{\mathbf{#1}}}

\title{On the adaptation of recurrent neural networks for system identification
\thanks{\textit{Corresponding author}: 
M. Forgione.} 
}

\author{
  Marco Forgione \\
  IDSIA\thanks{Dalle Molle Institute for Artificial Intelligence, IDSIA USI-SUPSI, Via la Santa 1, CH-6962 Lugano-Viganello, Switzerland.} \\
  Lugano\\
  \texttt{marco.forgione@supsi.ch} \\
   \And
  Aneri Muni \\
  NNAISENSE\thanks{NNAISENSE SA, Piazza Molino Nuovo 17, CH-6900 Lugano, Switzerland.} \\
  Lugano\\
  \texttt{aneri.muni@nnaisense.com} \\
  \And
  Dario Piga \\
  IDSIA \\
  Lugano\\
  \texttt{dario.piga@supsi.ch} \\
  \And
  Marco Gallieri \\
  NNAISENSE \\
  Lugano\\
  \texttt{marco.gallieri@nnaisense.com} \\
}

\begin{document}
\maketitle

\begin{abstract}
This paper presents a transfer learning approach which enables fast and efficient adaptation of Recurrent Neural Network (RNN) models of dynamical systems. A \emph{nominal} RNN model is first identified using available measurements. The system dynamics are then assumed to change, leading to an unacceptable degradation of the nominal model performance  on the \emph{perturbed} system. To cope with the  mismatch, the model is augmented  with an additive correction term trained on fresh data from the new dynamic regime. The correction term is learned through a Jacobian Feature Regression (JFR) method defined in terms of the features spanned by the model's Jacobian with respect to its nominal parameters.
A non-parametric view of the approach is also proposed, which extends recent work on Gaussian Process (GP) with Neural Tangent Kernel (NTK-GP) to the RNN case (RNTK-GP). This can be more efficient for very large networks or when only few data points are available. Implementation aspects for fast and efficient computation of the correction term, as well as the initial state estimation for the RNN model are described. Numerical examples show the effectiveness of the proposed methodology in presence of significant system variations. 
\end{abstract}


\section{Introduction}
\label{sec:intro}
In the Deep Learning (DL) field, expressive model structures are defined in terms of 
complex compositions of simple linear/non-linear building blocks \cite{schmidhuber2015deep}. Furthermore, model learning is performed conveniently using general gradient-based optimization, and leveraging  Automatic Differentiation (AD) for gradient computations \cite{baydin2017automatic}. Nowadays, specialized hardware and 
high-quality software tools for DL are available \cite{paszke2017automatic}, easing the practitioner's work to a large extent.

\begin{figure}[bt]
    \centering
    \includegraphics[width=0.70\textwidth]{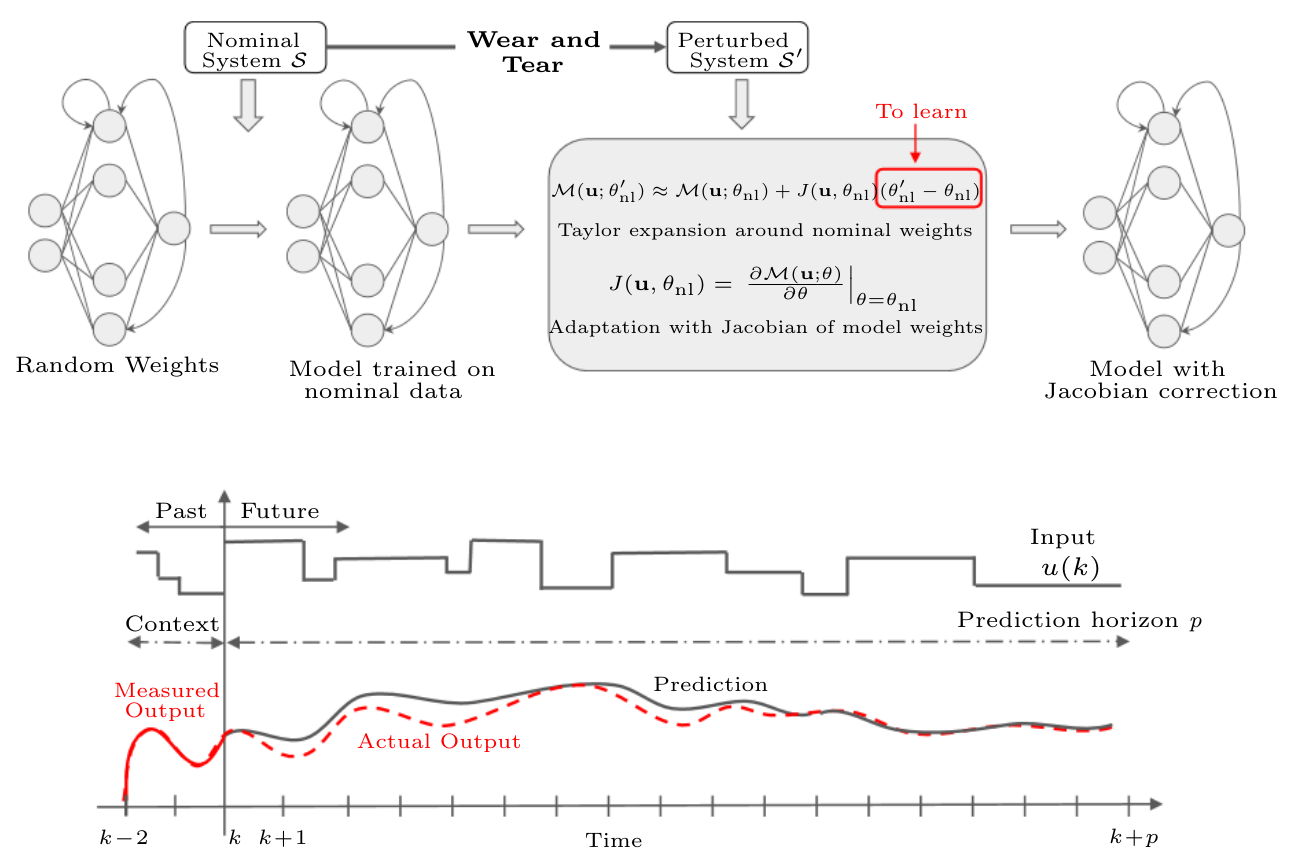}
    \caption{\textbf{(Top)} Nominal RNN model training and JFR adaptation with Jacobian features. \textbf{(Bottom)} Sequence handling for initialization of the RNN state.} 
    \label{fig:RNN-overview}
\end{figure}

In recent years, more and more research aimed at exploiting modern Deep Learning tools and concepts in System Identification (SI) has been pursued. Certain contributions are aimed at adapting existing DL architectures and training algorithms to the specific needs of SI. For instance, \cite{beintema2021nonlinear, forgione2021continuous} propose different approaches to  deal with the initial conditions in the training of Recurrent Neural Networks (RNN), while in \cite{andersson2019deep} 1D Convolutional Neural Networks are specialized and tested on SI tasks. Other contributions propose new DL-based model structures explicitly conceived for SI purposes. 
For instance, \cite{forgione2021dynonet} introduces a novel neural architecture which includes linear transfer
functions as elementary building blocks, while \cite{iacob2021deep} proposes model structures based on Koopman operator theory \cite{mauroy2020koopman}, and  lift non-linear state-space dynamics to a higher-dimensional space where the dynamics are linear. The mapping from the original to the higher-dimensional space is learned using DL tools.

DL-based models for SI have shown to deliver state-of-the-art performance on standard benchmarks, see \cite{iacob2021deep, mavkov2020integrated}.
In practice, however, the dynamics of real-world systems often vary over time due to, e.g., aging, changing configurations, and different external conditions. 
An open challenge is thus to \emph{adapt} the identified DL-based dynamical models over time.
In the \emph{systems and control} field, different model adaptation techniques have been introduced and successfully tested in practice. 
Joint estimation of states and (slowly-varying) model 
parameters has been tackled by enriching the model structure with a stochastic description of parameter evolution over time, and applying standard filtering techniques. Non-linear extension of the Kalman Filter (KF) such as the Unscented Kalman Filter (UKF)  \cite{pozzoli2020tustin} have been considered to obtain an (approximately) optimal estimate of state and parameters.

For models which are linear in the parameters, Bayesian Linear Regression (BLR) and \emph{Ridge} regression \cite{hastie01statisticallearning} as well as Recursive Least Squares (RLS) \cite{Ljung} enable  closed-form, even on-line estimation of the model parameters. For models that are not linear, on the other hand, the computational burden may become an issue. In \cite{pozzoli2020tustin}, the UKF has been used to adapt (online) the last layer of a particular neural architecture. This  entails the evaluation of the \emph{forward pass} two times the number of states and parameters, plus the inversion of a large matrix. The procedure is repeated as each new measurement becomes available and carries over an approximate prior. 

In this paper, we instead propose an approach to compute a local linear (with respect to its parameters) correction of the model dynamics. We use a Recurrent Neural Network nominal model and evaluate the linearization on entire sequences from the perturbed system in order to produce (state-dependent) Jacobian features that are used to perform JFR, and thus to update the model  using efficient least-squares methods. 
Our approach is inspired by the recent transfer learning literature  \cite{maddox2021fast}, where a first-order Taylor expansion of a nominal feed-forward neural network (modeling a static process) is introduced for model updating. With respect to \cite{maddox2021fast}, the contribution of the current paper is threefold:
\begin{itemize}
    \item we extend the methodology of \cite{maddox2021fast} to dynamical systems modeled as RNNs. In particular, we introduce {specialized} and computationally efficient derivations to obtain the RNN Jacobian features (needed for model updating) through \emph{recursive}  operations;  
    \item we introduce an approach to initialize the RNN state based on past data, so that an estimate of the initial state is not needed on top of the parameter estimation;  
    \item we extend the non-parametric view of the approach in \cite{maddox2021fast} by using the RNN Jacobian features to define a Recurrent Neural Tangent Kernel (RNTK) to be employed in a Gaussian Process (GP) framework. 
    \end{itemize}

The rest of this paper is organized as follows. Section \ref{sec:problem_formulation} is devoted to the problem formulation. The proposed approaches for nominal model training with initial state estimation  and 
transfer learning  are presented in Section~\ref{sec:trainadapt}. Details for efficient implementation are discussed in Section~\ref{sec:implementation}, along with analysis of  the required computational and memory costs. The effectiveness of the proposed methodology is showcased in Section \ref{sec:experiments} on numerical examples from the chemical and electrical domains.

\section{Problem formulation} 
\label{sec:problem_formulation}

Let us consider a discrete-time \emph{dynamical system} $\sys$ taking input values $u \in \R^\nin$ and generating output values $y \in \R^\ny$. 
At a given discrete-time step $k$, the system output possibly depends on all the previous input samples, \emph{i.e.},
\begin{equation}
    y_{k} = \sys(u_{k}, u_{k-1}, \dots, u_{0}).
\end{equation}
Let $\M$ be a non-linear  \emph{dynamical model structure} sufficiently expressive to describe the system $\sys$:
\begin{equation}
\hat y_{k} = \M(u_{k}, u_{k-1}, \dots, u_{0}; \;\theta),
\end{equation}
where $\theta \in \R^{\npar}$ is a \emph{parameter vector} to be determined. 

We use the bold-face symbols $\tvec{u}$ and $\tvec{y}$ to denote the sequences of $\nsamp$ input and output samples collected during an experiment. 
For the mathematical derivations, $\tvec{u}$ and $\tvec{y}$ may be interpreted as real-valued vectors of size $\nsamp$.
The true system  and the model dynamics will be written compactly in vector notation as $\sys(\tvec{u})$
and $\M(\tvec{u};\; \theta)$, respectively.

In this paper, we assume that the
model admits the following state-space representation:
\begin{subequations}
\label{eq:statespace_model}
\begin{align}
    x_{k+1} &=  \F(x_k, u_k; \theta_{\F}) \label{eq:statespace_model_state} \\
    \hat y_k   &= \G(x_k, u_k; \theta_{\G}) \label{eq:statespace_model_out},
\end{align}
\end{subequations}
where $x_{k} \in \R^{\nx}$ is the \emph{state vector} at time $k$, and $\F, \G$ are the state-update and output mappings, respectively, both parametrized by $\theta=\{\theta_\F,\theta_\G\}$. In particular, $\F$ and $\G$ in this paper are neural networks, and the overall model \eqref{eq:statespace_model} is a Recurrent Neural Network (RNN). 

We denote by $\ops(\F)$ and $\ops(\G)$ the computational cost required to simulate $\F$ and $\G$, respectively and we assume that $\ops(\F)$ and $\ops(\G)$ are both $\mathcal{O}(\npar)$.
The computational cost of simulating the RNN over $\nsamp$ time
steps is thus $\ops(\M) = \nsamp \left(\ops(\F) + \ops(\G)\right)$, 
or simply $\mathcal{O}(\nsamp \npar)$.


The system  $\sys$ operates initially in its \emph{nominal} configuration. In this configuration, the training dataset $\D_\tr = \{\tvec{u}_\tr, \tvec{y}_\tr\}$ and the test dataset 
$\D_\tst = \{\tvec{u}_\tst, \tvec{y}_\tst\}$ are collected. These datasets are used to 
learn the parameter vector $\theta_\nominal$ of a \emph{nominal} non-linear model 
and to asses its performance, respectively.

After some time, the system dynamics changes due to, e.g., aging or different external conditions.
We denote by $\sys'$ the system in this \emph{perturbed} configuration and we assume that the nominal model
(previously trained on $\D_\tr$) no longer describes of the behavior of $\sys'$ with sufficient accuracy.

While we could re-run the full non-linear model training procedure on a new training dataset collected in the perturbed configuration, we would like to exploit the previously trained model and \emph{adapt} it to obtain a description of the perturbed system in a more efficient manner.  In this spirit, we collect a \emph{transfer dataset} $\D_\transf = \{\tvec{u}_\transf, \tvec{y}_\transf\}$ and an \emph{evaluation dataset} $\D_\eval = \{\tvec{u}_\eval, \tvec{y}_\eval\}$ from $\sys'$. The purposes of these datasets are to adapt the nominal model to the perturbed configuration and to evaluate its performance, respectively.

In general, the model adaptation procedure should be less computational intensive and/or require less data than a (trivial) repetition of the training procedure from scratch.

In the following sections, derivations are shown for the single-input-single-output case for notational simplicity (i.e. $n_u = n_{y} = 1$). Extension to the multi-input-multi-output case is trivial, and considered in the Continuous-flow Stirred-Tank Reactor example presented in Section~\ref{sec:experiments}.

\section{Model training and adaptation} 
\label{sec:trainadapt}
\subsection{Nominal model training}
\label{sec:nominal_training}
The parameter vector $\theta_\nominal$ of the nominal model may be selected by minimizing a {regression loss} such as the \emph{mean square error} $\mathrm{MSE}(\D_{\tr}, \theta)$ with respect to $\theta$ on the training dataset
$\D_{\tr}$:
\begin{equation}
    \label{eq:training_loss}
    \theta_\nominal = \arg \min_{\theta \in \R^\npar} 
    \overbrace{\frac{1}{N}\norm{\tvec{y}_\tr - \M(\tvec{u}_\tr; \theta) }^2}^{=\mathrm{MSE}(\D_{\tr}, \theta)}.
\end{equation}
The loss is minimized through iterative gradient-based optimization algorithms, such as plain gradient descent or modern variants like Adam \cite{kingma2014adam}. For the required derivatives computation, standard reverse-mode Automatic Differentiation (AD) algorithms and software are used \cite{paszke2017automatic}. 
Accordingly, the number of operations required to compute the gradient of the loss w.r.t. $\theta$ is $k \ops(\M)$, where $k$ is a constant guaranteed to be $k < 6$, see \cite{baydin2017automatic}. Thus, gradient computation (which is the 
most expensive operation for each iteration of gradient-based optimization) for a RNN model has cost $\mathcal{O}(\nsamp \npar)$. 
Of course, a certain number $n_{\rm iter}$ of optimization iterations (hard to determine a priori) is needed
to reduce the training loss to a reasonably low value. The overall computational cost of solving \eqref{eq:training_loss} is thus $\mathcal{O}(n_{\rm iter} \nsamp \npar)$.
We will not discuss the nominal model training in more details, as standard practice is followed.

\subsection{RNN initial state estimation}
\label{OpenLSTM}
The initial state of a RNN, generally set to zero, can affect the transient predictions from the model. Capturing the first steps accurately is actually very important, for instance in applications such as Model Predictive Control (MPC) \cite{Maciejowski_book}.   Estimating the initial state is addressed in~\cite{Armenio2019} using  a past window of both measurements and outputs, with guarantees provided for Echo State Networks (ESN). We extend this idea (without explicit guarantees) to a more general class of RNNs, where the predicted output $\hat{y}_k$ is part of the state:  
\begin{subequations}
\label{eq:statespace_model_factored}
\begin{align}
    x_{k+1} &=  {\F}(x_k, \hat{y}_k, u_k; \theta_\F)\\
    \hat{y}_{k+1}   &= {\G}(x_{k+1}, \hat{y}_k, u_k; \theta_\G).
\end{align}
\end{subequations}

While predictions are performed by (\ref{eq:statespace_model_factored}),  the initial state is estimated using a window of past data (context), $\{y_{k-1},\dots,y_{k-N_c}, u_{k-1},\dots, u_{k-N_c}\}$, where the \emph{measured} outpus $y_{k-i}$ are sequentially fed to the model instead of the predicted ones for $N_c$ steps,  and $x_{k-N_c}$ is set to zero. In other words, the state estimation is performed by opening the output prediction loop for the first for $N_c$ steps:
\begin{subequations}
\label{eq:statespace_model_estimator}
\begin{align}
    & x_{k+1} =  {\F}(x_k, y_k, u_k; \theta_\F)\\
    & \hat{y}_{k+1}   = {\G}(x_{k+1}, {y}_k, u_k; \theta_\G),\\
    &\text{for $k=0,1,\dots, N_c\!-\!1$.}
\end{align}
\end{subequations} 
Since the state estimator is the model itself, we train by means of a joint backward pass through both estimation and forecasting. Sequence are split as in Figure \ref{fig:RNN-overview} (bottom chart). 




\subsection{Model adaptation}
\label{subsec:BLR}
Suppose the dynamical system $\sys$ changes over time, and the performance of 
the nominal trained model $\M(\cdot; \theta_\nominal)$ on the new dynamics $\sys'$ drops to an unacceptable level.
Following the reasoning in \cite{maddox2021fast}, we exploit
the Jacobian $J(\tvec{u}, \theta_\nominal) \in \R^{\nsamp \times \npar}$ of the nominal model $\M(\cdot; \theta_\nominal)$ with respect to 
the parameter vector $\theta$:
\begin{equation}
    J(\tvec{u}, \theta_\nominal) = \left . \frac{\partial \M(\tvec{u}; \theta)}{\partial \theta} \right |_{\theta = \theta_\nominal}
\end{equation}
to define an \emph{adaptation} of the nominal model:
\begin{equation}
\label{eq:model_adaptation}
\M_\lin(\tvec{u}; \theta_{\nominal}, \theta_\lin) = \M({\tvec{u}}; \theta_\nominal) + J(\tvec{u}, \theta_\nominal) \theta_\lin,
\end{equation}
which depends \emph{linearly} on a new parameter vector $\theta_\lin \in \R^{n_{\theta}}$, to be determined. 
The rationale behind this approach is the following---the perturbed system dynamics $\sys'$ may be described 
by a model $\M({\tvec{u}}; \theta_\nominal')$ with a parameter $\theta_\nominal'$ slightly different from $\theta_\nominal$.
Then, $\M({\tvec{u}}; \theta_\nominal')$ may be approximated through a first-order Taylor expansion centered at  
the nominal parameter $\theta_\nominal$: 
\begin{equation*}
\M({\tvec{u}}; \theta_\nominal') \approx \M({\tvec{u}}; \theta_\nominal) + J(\tvec{u}, \theta_\nominal) 
(\theta_\nominal' - \theta_\nominal)
,
\end{equation*}
which corresponds indeed to the model structure \eqref{eq:model_adaptation} by setting $\theta_\lin = \theta_\nominal' - \theta_\nominal$.

In the following, for notation simplicity, we ignore the nominal non-linear contribution $\M({\tvec{u}}; \theta_\nominal)$ at the right-hand side of model \eqref{eq:model_adaptation} and only retain the linear adaptation term $J(\tvec{u}, \theta_\nominal) \theta_\lin$. 
We estimate the parameters $\theta_\lin$ of the linear adaptation term through a Jacobian Feature Regression (JFR), according to 
the probabilistic model:
\begin{subequations}
\label{eq:bayesmodel}
\begin{align}
    \tvec{y} &= 
    J(\tvec{u}, \theta_\nominal)\theta_\lin + \tvec{e}, \qquad  \tvec{e} \sim \N(0, \sigma^2I_N)\\
    \theta_\lin &\sim \N(0, I_{\npar}).
\end{align}
\end{subequations}
From \eqref{eq:bayesmodel}, the \emph{posterior} distribution of $\theta_\lin$ conditioned on the transfer dataset $\D_\transf$ is Gaussian: 
\begin{subequations}
\label{eq:meanvar}
\begin{equation}
    \theta_\lin | \D_\transf \sim \N(\bar{\theta}, \Sigma_{\theta_\lin})
\end{equation}
with mean
\begin{equation}
    \label{eq:thetalinmean}
    \bar{\theta} = (J^\top_\transf J_\transf + \sigma^2 I_\npar)^{-1}J^\top_\transf \tvec{y}_\transf 
\end{equation}
and covariance matrix:
\begin{equation}
\label{eq:thetalincovar}
\Sigma_{\theta} = \sigma^2(J_\transf^\top J_\transf + \sigma^2 I_\npar)^{-1},
\end{equation}
\end{subequations}
where $J_\transf = J(\tvec{u}_\transf, \theta_\nominal)$.

The posterior distribution of $\theta_\lin$ given by \eqref{eq:meanvar} may be used to infer the output distribution given a new input sequence $\tvec{u}_*$ according to:

\begin{equation}
    \label{eq:posterior_parspace_simple}
   \tvec{y}_* | \D_\transf, \tvec{u}_* \sim \N(\;\overbrace{J_* \bar{\theta}}^{=\tvec{\bar{y}}_*},\; J_* \Sigma_{\theta} J_*^\top\;),
\end{equation}
where $J_* = J(\tvec{u}_*, \theta_\nominal) = \left . \frac{\partial \M(\tvec{u_*}; \theta)}{\partial \theta} \right |_{\theta = \theta_\nominal}$.

\subsection{Interpretation in function space}
\label{sec:funspace_interp}
Similar to \cite{maddox2021fast}, the RNN-feature JFR introduced above can be equivalently formulated as a Gaussian Process (GP). In this work, the GP is parametrized with the finite-dimensional \emph{Recurrent} Neural Tangent Kernel (RNTK) resulting from the RNN trained on nominal data. Using \eqref{eq:meanvar} and \eqref{eq:posterior_parspace_simple}, we have the following predictive posterior output distribution:
\begin{equation}
\label{eq:posterior_parspace_full}
 \tvec{y}_* | \tvec{u}_*, \D_\transf \sim \N\big(J_* (J^\top_\transf J_\transf + \sigma^2 I_\npar)^{-1}J^\top_\transf \tvec{y}_\transf, 
 \sigma^2 J_* (J_\transf^\top J_\transf + \sigma^2 I_\npar)^{-1} J_*^\top\big).
\end{equation}
Using the matrix inversion lemma and following the derivations in  \cite[Chapter~1]{rasmussen2006gaussian}, the predictive posterior distribution may be equivalently written as in \cite[Equation 3]{maddox2021fast}:
\begin{equation}
\label{eq:posterior_funspace}
 \tvec{y}_* | \tvec{u}_*, \D_\transf \sim \N(
 J_* J_\transf^\top(J_\transf J_\transf^\top + \sigma^2 I_\nsamp)^{-1}\tvec{y}_\transf, 
 \sigma^2 J_*(I_\nsamp - J_\transf^\top(J_\transf J_\transf^\top + \sigma^2 I_{\npar})^{-1} 
 J_\transf)J_*^\top,
\end{equation}
with the difference that, instead of computing $J_\transf$ for a deep network with point predictions, this is now computed for a RNN sequence-to-sequence task.

\section{Implementation}
\label{sec:implementation}

The computational cost  of inferring  the posterior mean $\bar{\tvec{y}}_*$ of the output $\tvec{y}_*$ is analyzed in this section, both in parameter and in function space.

\subsection{Parameter-space inference}\label{sec:parspace_inferece}
The adaptation entails two steps: 1) Run the RNN on a fresh, small dataset and compute the posterior parameter mean $\bar{\theta}$; 2) Run the RNN on new data, compute the predictive posterior mean based on $\bar{\theta}$ and the current Jacobian. The scheme is depicted at the top of Figure \ref{fig:RNN-overview}.

For parameter-space inference, the main problem to be tackled is the computation of the posterior parameter mean $\bar{\theta}$. Once $\bar{\theta}$ is available, the posterior 
output mean    is given by:  $$\bar{\tvec{y}}_* =   J_* \bar{\theta},$$ see eq.~\eqref{eq:posterior_parspace_simple}. 
The \emph{Jacobian-vector product} $ J_* \bar{\theta}$ can be obtained efficiently exploiting the same automatic differentiation tools already used in \ref{sec:nominal_training} to compute (for each iteration of gradient-based optimization) the derivatives of the training loss \eqref{eq:training_loss}, thus with a number of operations $\mathcal{O}(\nsamp \npar)$, and without explicitly constructing the Jacobian matrix $J_*$. 
The reader is referred to \cite{balestriero2021fast, baydin2017automatic} for the details about Jacobian-vector product computations through  AD. 

In the following paragraphs, we discuss three different implementation approaches to compute the posterior parameter mean $\bar{\theta}$, along with a computationally efficient derivation to obtain the Jacobian $J_\transf$. 


\subsubsection{Offline implementation}
In the offline implementation approach, the estimation of $\bar \theta$ is carried out once the complete transfer 
dataset $\D_\transf$ is available.
The main challenge is to compute the full Jacobian matrix $J_\transf$. Once available,    the system of  linear equations \eqref{eq:thetalinmean} is constructed and solved to obtain $\bar{\theta}$. Details are provided below.\\

\noindent \textbf{\emph{Computation of the Jacobian matrix $\bm{J_\transf}$.}} A straightforward approach to compute the Jacobian $J_\transf \in \R^{\nsamp \times \npar}$ consists in invoking $N$ independent reverse-mode automatic differentiation (back-propagation) operations to construct the $N$ rows of the matrix. The computational cost of this na\"ive approach is $\mathcal{O}(N^2 \npar)$.

In this paper, we introduce an alternative approach tailored to the RNN model \eqref{eq:statespace_model}, based on a recursive construction of the Jacobian matrix, which exploits forward sensitivity equations \cite{rabitz1983sensitivity}. This approach has a computational cost $\mathcal{O}(N (\nx + \ny) \npar)$. 

Let us introduce the \emph{state sensitivities} $s_{k} = \frac{\partial x_{k}}{\partial \theta}\in R^{\nx \times \npar}$.
By taking the derivatives of the left- and right-hand side of~\eqref{eq:statespace_model_state} w.r.t. the model parameters $\theta$, we obtain a recursive equation describing the evolution of $s_k$:
\begin{subequations}
\label{eq:sens}
\begin{equation}
    \label{eq:sens_x}
    s_{k+1} = J^{fx}_k s_k + J^{f\theta}_k,
\end{equation}
where $J^{fx}_k \in \R^{\nx \times \nx}$ and $J^{f\theta}_k \in \R^{\nx \times \npar}$ are the Jacobians of  $\mathcal{F}(x_k, u_k; \theta)$ w.r.t. $x_k$ and $\theta$, respectively.

Let us now take the derivative of \eqref{eq:statespace_model_out} w.r.t. $\theta$, which corresponds by definition to the $k$-th row of $J_{\transf}$:
\begin{align}
\label{eq:sens_y}
    \frac{\partial y_k}{\partial \theta} = J^{gx}_{k} s_k + J^{g\theta}_{k},
\end{align}
where $J^{gx}_{k} \in \R^{\ny \times \nx}$ and $J^{g\theta}_{k} \in \R^{ny \times \npar}$ are the Jacobians of  $\mathcal{G}(x_{k}, u_{k}; \theta)$ w.r.t. $x_{k}$ and $\theta$, respectively.
\end{subequations}

The Jacobians $J^{fx}_{k}$ and $J^{f\theta}_{k}$ can be obtained through $\nx$ back-propagation operations through $\F$, thus at cost $\mathcal{O}(\nx \npar)$. Similarly, $J^{gx}_{k}$ and $J^{g\theta}_{k}$ can be obtained through $n_y$ back-propagation operations through $\G$ at cost $\ny \npar$.
Overall, the computational cost of obtaining $\frac{\partial y_k}{\partial \theta}$ in ~\eqref{eq:sens_y} (given the previous sensitivity $s_{k-1}$) is $\mathcal{O}((\nx + n_y)\npar)$. Thus, $J_{\transf}$ is obtained at a cost $\mathcal{O}(\nsamp (\nx + n_y)\npar)$.\\

\noindent \textbf{\emph{Solution of the linear system.}}
The linear system to be solved in order to obtain $\bar{\theta}$ in \eqref{eq:thetalinmean} is:
\begin{equation}
\label{eq:linear_system}
\overbrace{(J^\top_\transf J_\transf + \sigma^2 I_\npar)}^{=A}\bar{\theta} = \overbrace{J^\top_\transf \tvec{y}_\transf}^{=b},
\end{equation}
where the system matrix $A \in \R^{\npar \times \npar}$ is symmetric and positive definite.
The standard approach to solve such a linear system involves application of the Cholesky factorization \cite[Appendix~A]{nocedal2006numerical}, which requires approximately $\frac{\npar^3}{3}$ operations. 
Thus, the computational cost is $\mathcal{O}(\npar^3)$, while the memory cost is $\mathcal{O}(\npar^2)$, since the system matrix $A$ has to be stored explicitly into memory.

\subsubsection{Online implementation}
Thanks to  the linear-in-the-parameters structure of the adapted model $\M_\lin$
 in~\eqref{eq:model_adaptation},  the parameter vector $\bar{\theta}$ may be estimated in real time by applying the Recursive Least Squares (RLS) algorithm \cite[Chapter~11]{ljung:1999system}, as soon
as a new sample is measured.

Implementation of RLS requires, at each time step, simple linear algebra operations
(sums and multiplications) 
on matrices of size $\npar \times \npar$ and vectors of size $\npar$. Furthermore, the Jacobian feature $\frac{\partial y_k}{\partial \theta}$ (which is the regressor required in RLS at time step $k$)  is  recursively obtained by  exploiting the forward sensitivities equations  \eqref{eq:sens},  thus with computational cost $\mathcal{O}((\nx + n_y)\npar)$.

\subsubsection{Limited-memory implementation}
\label{sec:LM-BLR}
In certain cases, the number of parameters $\npar$ may be too large to adopt the approaches discussed above. 
In particular, the matrices $A$ in \eqref{eq:linear_system} and the matrices to be updated in RLS, all having size $\npar \times \npar$, may exceed the computing device's available memory for large-scale RNN  models. 

In this case, as already mentioned at the beginning   of Section~\ref{sec:parspace_inferece}, for a given vector $v$ of compatible size, it is still possible to compute the Jacobian-vector product  $J_\transf v$ (and actually also the \emph{transposed} Jacobian-vector product  $J_\transf^\top v$) using the very same machinery of a gradient evaluation through AD without explicitly computing and storing the Jacobian matrix $J_\transf$, at computational cost  $\mathcal{O}(\nsamp \npar)$.

For limited-memory computation of $\bar \theta$, it is then convenient to reformulate  \eqref{eq:thetalinmean} as the solution of the optimization problem:
\begin{equation}
    \label{eq:thetalinmean_opt}
    \bar{\theta} = \arg \min_{\theta} 
    \overbrace{\norm{\tvec{y}_\transf - J_\transf \theta}^2 + \sigma^2 \theta^\top \theta}^{=\mathcal{L}(\D_\transf, \theta)}.
\end{equation}
Evaluation of the loss $\mathcal{L}(\D_\transf, \theta)$ in \eqref{eq:thetalinmean_opt}  may be performed by computing the Jacobian-vector product $J_\transf \theta$ through AD.
Furthermore, the gradients $\frac{\partial \mathcal{L}(\D_\transf, \theta)}{\partial \theta}$ of the loss $\mathcal{L}$ w.r.t. $\theta$ is:
\begin{equation*}
\frac{\partial \mathcal{L}(\D_\transf, \theta)}{\partial \theta} = 2\ J_\transf^\top (J_\transf \theta - \tvec{y}_\transf) + 2\ \sigma^2\theta,
\end{equation*}
and it can be obtained through the additional transposed Jacobian-vector product $J_\transf^\top (J_\transf \theta - \tvec{y}_\transf)$.
Thus, an approximate numerical solution to \eqref{eq:thetalinmean_opt} may be obtained by means of \emph{iterative} gradient-based optimization techniques, where each iteration has computational cost $\mathcal{O}(\nsamp \npar)$ and the same memory cost of back-propagation through the RNN model. 

\subsection{Function-space inference}
\label{sec:GP_func}
As shown in \cite{maddox2021fast}, the function-space formulation \eqref{eq:posterior_funspace} can leverage the fast Jacobian-vector products, similar to our parametric limited-memory approach.  The final model correction is computed by means of the conjugate gradient method \cite[Chapter~5]{nocedal2006numerical}, for an overall effort of $\mathcal{O}(N^2)$ for each iteration of the conjugate gradient. Thus, the approach can become favourable for very large models and small datasets, provided that convergence occurs within few iterations.   

The implementation of \cite{maddox2021fast} is used for the RNTK GP. The network state is initialized first using the nominal model, then a NTK GP is created which accepts 2D tensors as the network input and output. For compatibility, given the estimated RNN inital state, the model is wrapped in an interface that allows the IO sequence length to be collapsed over the batch size. Note that the  contribution of past states and inputs is accounted for when computing the backward passes used for Jacobian-vector products. In this respect, the proposed RNTK extends the approach of \cite{maddox2021fast}.

\section{Examples} 
\label{sec:experiments}
Two demonstrators are presented, respectively, from the chemical and electrical domain. 
For both examples, we have defined a nominal and a perturbed data-generating system.
The nominal system is used to generate the training and the test datasets, while the perturbed system 
is used to generate the transfer and the evaluation datasets. First, a nominal model is estimated on 
the training dataset using standard gradient-based approaches, and its performance is evaluated on the test dataset. 
Then, the nominal model is adapted to the perturbed configuration using data from the transfer dataset, following the approach described in Section \ref{sec:implementation}.
On the evaluation dataset, we can finally measure the performance of the adapted model in the perturbed configuration and compare it to the 
performance of the nominal model.

The performance of the estimated nominal and adapted model is  assessed 
through  the    $R^2$  \emph{performance index}:
\begin{equation}
    R^{2}(\tvec{y}, \hat{\tvec{y}}) = 1 - \frac{\norm{\tvec{y} - \hat{\tvec{y}} }^2 } {{\norm{\tvec{y} - \overline{\tvec{y}}}^2}},
\end{equation}
where $\overline{\tvec{y}}$ denotes the  mean of the measured output $\tvec{y}$, and $\hat{\tvec{y}}$ is the predicted output from the model. 

The developed software is based on the PyTorch DL framework \cite{paszke2017automatic}. 
The implementation for function-space inference builds upon the codes of the paper \cite{maddox2021fast}, 
adapted to handle a RNN nominal model. Instead, the parameter-space methods have been developed 
from scratch. All our codes are available on the GitHub repository \url{https://github.com/forgi86/RNN-adaptation} and allow full reproduction of the results in this paper.

We run all the experiments PC equipped with an Intel i7-8550U CPU and 16~GB of RAM.

\subsection{Continuous-flow Stirred-Tank Reactor (CSTR)}
As a first case study, we consider the CSTR system~\cite{CSTR_textbook}, with continuous-time dynamics described by the following \emph{ordinary differential equations}:
\begin{align}
\dot C_A &= 
q (C_{A}^0 - C_A) - k^1 C_A + k^4 C_R \nonumber \\
\dot C_R &= q(1 - C_{A}^0 - C_R) \!+\! k^1 C_R \!+\! k^3 (1 - C_A - C_R) \nonumber \\ & \quad \!-\! (k^2 + k^4) C_R, \label{eqn:CSTR}
\end{align}
with inputs  $T$ and  $q$ denoting,  the temperature and the flow rate, respectively;  output concentrations $C_A$ and $C_R$;  and coefficient   $k^i$ defined as  
\begin{equation}
    k^i = k_{0}^{i} \exp\left(-E^i \left(\frac{1}{T} - 1\right)\right), \qquad i=1,2,3,4.
\end{equation}
The nominal and perturbed system parameters are reported in Table~\ref{tab:CSTR_coefficients}.

\begin{table}[b]
\centering
\caption{CSTR example: nominal and perturbed system parameters.}
\begin{tabular}{l | c  c  c  c c c c c c}
 \hspace*{-0.15cm} \textbf{System} \! \!  \hspace*{-0.2cm}& \multicolumn{8}{c}{\textbf{Parameters}} \\
                &\! \! \hspace*{-0.33cm} $C_{A}^0$ \! \! \hspace*{-0.2cm} &\! \! \hspace*{-0.33cm} $k_{0}^{1}$ \! \! & \! \! \hspace*{-0.33cm} $k_{0}^{2}$ & \! \! \hspace*{-0.33cm} $k_{0}^{3}$ & \! \! \hspace*{-0.33cm} $k_{0}^{4}$ & \! \! \hspace*{-0.33cm} $E^{1}$ & \! \! \hspace*{-0.33cm} $E^{2}$ & \! \! \hspace*{-0.33cm} $E^{3}$ & $E^{4}$ \\ 
\hline
\hspace*{-0.15cm} nominal   \!\!  \hspace*{-0.2cm}    &\! \! \hspace*{-0.33cm} 0.8 \! \! \hspace*{-0.2cm}  & \! \! \hspace*{-0.33cm} 1 \! \!  & \! \! \hspace*{-0.33cm} 0.7& \! \! \hspace*{-0.33cm} 0.1 & \! \! \hspace*{-0.33cm} 0.006 & \! \! \hspace*{-0.33cm}  8.33 & \! \! \hspace*{-0.33cm} 10 & \! \! \hspace*{-0.33cm} 50 & \! \! \hspace*{-0.33cm} 83.3 \\
\hspace*{-0.15cm} perturbed \!\!  \hspace*{-0.2cm}    &\! \! \hspace*{-0.33cm} 0.8 \! \! \hspace*{-0.2cm} & \! \! \hspace*{-0.33cm} 1 \! \!  & \! \! \hspace*{-0.33cm} 0.7& \! \! \hspace*{-0.33cm} 0.1 & \! \! \hspace*{-0.33cm} 0.006 & \! \! \hspace*{-0.33cm} 7.33 & \! \! \hspace*{-0.33cm}  9 & \! \! \hspace*{-0.33cm} 60 & \! \! \hspace*{-0.33cm} 93.3 \\
\hline
\end{tabular}
\label{tab:CSTR_coefficients}
\end{table}
The model~\eqref{eqn:CSTR}  is discretized with sampling time $T_s = 0.1~\rm{s}$. We collect a training dataset consisting of 64 sequences containing 256 samples, and a test, transfer, and evaluation dataset each consisting of a single sequence of length 1024. In all datasets, the input temperature $T$  is a triangular wave and the feed rate $q$ is a step signal with random amplitude values.

We model the system with an LSTM network \cite{hochreiter1997a,greff_lstm_2017} having two inputs; one hidden layer with 16 units; and 2 output units. The nominal LSTM model parameters are obtained by minimizing the mean square simulation error on the training dataset using standard gradient descent techniques. Moreover, to assist with the tracking of the signal transient, we initialize the LSTM with a context, as illustrated in Section \ref{OpenLSTM}. 

The achieved $R^2$ indexes in estimating the two outputs $C_A$ and $C_R$ are  reported in Table \ref{tab:CSTR_performance}, where we can observe that the performance of the nominal LSTM model is satisfactory on both the training and test datasets. However, the performance index drops significantly on the transfer and evaluation datasets. This shows that the nominal LSTM model cannot accurately describe the perturbed system dynamics.

We now utilize the transfer dataset to adapt the nominal model using the following three different approaches described in Sections~\ref{sec:trainadapt}  and~\ref{sec:implementation}: 
\begin{itemize}
    \item Jacobian Feature Regression (JFR) with standard offline implementation described in Section~\ref{subsec:BLR} 
    and based on  eqs.~\eqref{eq:sens} and~\eqref{eq:linear_system};
    \item Limited-memory Jacobian Feature Regression (LM-JFR) based on an iterative gradient-based minimization of \eqref{eq:thetalinmean_opt}, exploiting Jacobian-vector products for gradient computation, as described in Section~\ref{sec:LM-BLR};
    \item Function-space approach based on a Gaussian Process with  kernel designed from the Jacobian of the LSTM (GP-LSTM), as described in Sections~\ref{sec:funspace_interp} and~\ref{sec:GP_func}. 
\end{itemize}

\begin{figure}[bt]
    \centering
    \includegraphics[width=0.70\textwidth]{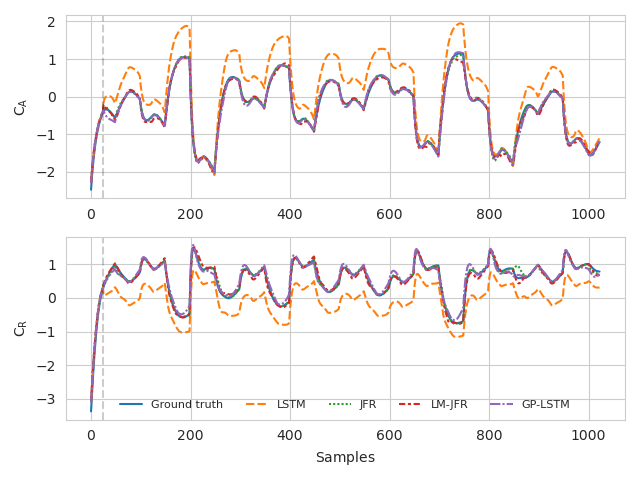}
    \caption{ CSTR example. Normalized output concentrations $C_A$ and $C_R$ on the evaluation dataset: ground truth (blue); nominal LSTM (orange);   Jacobian Feature Regression (green); Limited Memory Jacobian-Vector Product implementation (red); non-parametric GP-LSTM (purple). The gray vertical lines represent the context window consisting of the first 25 datapoints.} 
    \label{LSTM-compare1}
\end{figure}

 Fig.~\ref{LSTM-compare1} compares the (normalized) ground truth output with the nominal LSTM output and 
 the outputs of the JFR, LM-JFR, and GP-LSTM model adaptation approaches on the evaluation dataset. The 
 outputs of the three model adaptation approaches are shown to be equal (up to small numerical deviations). 
 The $R^2$ index of the nominal LSTM and of the adapted model (equivalent for the three approaches) is reported in Table~\ref{tab:CSTR_performance}. We observe that on the evaluation dataset the adapted model achieves excellent predictive performances, with an $R^2$ index above 0.99 for both output channels. 
 Conversely, the performance of the nominal LSTM decreases dramatically on the evaluation dataset ($R^2$ index of 0.5/-0.74 for the two output channels).
  
 The run time of the three model adaptation approaches is reported in Table~\ref{tab:CSTR_comp_time}, which show that, for this particular example, JFR turned out to be the fastest one.  
 
\begin{table}[tb]
\caption{CSTR example: Computational time  of performing transfer for proposed methods JFR, LM-JFR, GP-LSTM. }
\centering
\begin{tabular}{l | c }
\textbf{Model} & \textbf{Computational Time  (s)} \\
\hline
JFR          & 13.24  \\
LM-JFR       & 295.37  \\
GP-LSTM      & 1321.05 \\
\hline
\end{tabular}
\label{tab:CSTR_comp_time}
\end{table} 

\begin{table}[tb]
\caption{CSTR example: $R^2$ performance index of  nominal and  adapted model on the two output channel $C_A$/$C_R$ for training, test, transfer and evaluation  datasets. }
\centering
\begin{tabular}{l | c  c  c  c}
\multirow{2}{*}{\textbf{Model}} & \multicolumn{4}{c}{\textbf{Dataset}} \\
                & Train      & Test       & Transfer & Evaluation       \\ 
\hline
nominal         & 0.99/0.99       & 0.99/0.99        & 0.74/0.08      & 0.50/-0.74  \\
adapted         & -          & -          & 0.99/0.99     & 0.99/0.99  \\
\hline
\end{tabular}
\label{tab:CSTR_performance}
\end{table}

\subsection{Non-linear RLC}
We consider the non-linear RLC series circuit benchmark introduced in \cite{forgione2021continuous} and described by:
\begin{equation}
\label{eq:RLC_sys}
\begin{bmatrix}
\dot v_C\\
\dot i_L
\end{bmatrix} = 
\begin{bmatrix}
  0           & \tfrac{1}{C}\\
 \tfrac{-1}{L(i_L)} & \tfrac{-R}{L(i_L)}\\
\end{bmatrix}
\begin{bmatrix}
v_C\\
i_L
\end{bmatrix} +
\begin{bmatrix}
0\\
\tfrac{1}{L(i_L)}
\end{bmatrix} 
v_{in},
\end{equation}
where $v_{in}~\rm(V)$ is the input voltage; $v_C~\rm(V)$ is the capacitor voltage; and $i_L~\rm(A)$ is the inductor current. The resistance $R$ and capacitance $C$ are constant, while the inductance $L$ depends on $i_L$ according to:
\begin{equation*}
 L(i_L) = L_0\bigg[\bigg(\frac{0.9}{\pi}\arctan\big(-\!5(|i_L|-5\big)+0.5\bigg) + 0.1 \bigg]. 
\end{equation*}
The coefficients characterizing the nominal and the perturbed systems are reported in 
Table~\ref{tab:RLC_coefficients}. 
All datasets contain 2000 samples obtained by simulating the (nominal or perturbed) system with discretization step $T_s = 1$~$\mu$s. 
The input is a filtered white noise with bandwidth $80, 90, 100, 100~\text{kHz}$ and standard deviation $80, 70, 70, 70~\text{V}$ for the training, test, transfer, and evaluation datasets, respectively. 
In the training and transfer datasets, the output $v_C$ is corrupted by an additive white noise term. The corresponding signal-to-noise ratios are $20.0$ and $18.8$~dB, respectively.

By considering as state vector $x = [v_C\; i_L]^\top$ and input $u=v_{in}$, we adopt for this system the following nominal state-space model structure:
\begin{align*}
\dot x & = \mathcal{F}(x, u; \theta) \\
 y &= v_{C},
\end{align*}
where $\mathcal \F$ is a feed-forward neural network with three input units (corresponding to $v_C$, $i_L$, and $v_{in}$); a hidden layer with 64 linear units followed by tanh nonlinearity; and two linear output units corresponding to the components of the state equation to be learned. 

The model above is discretized with the forward-Euler method at sampling time $T_s = 1$~$\mu$s, resulting in a 
discrete-time RNN model. The parameters $\theta$ are then estimated on the training dataset using the  {truncated simulation error minimization} method described in~\cite{forgione2021continuous}. 
In particular, training is performed over 10000 iterations of gradient-based optimization on minibatches, each one containing
16 sequences of length 256 extracted from the training dataset. 

Model adaptation is performed in the parameter-space setting, through  the Bayesian Linear Regression approach discussed in the paper. 

\begin{figure}[ht]
    \centering
\includegraphics[width=0.70\textwidth]{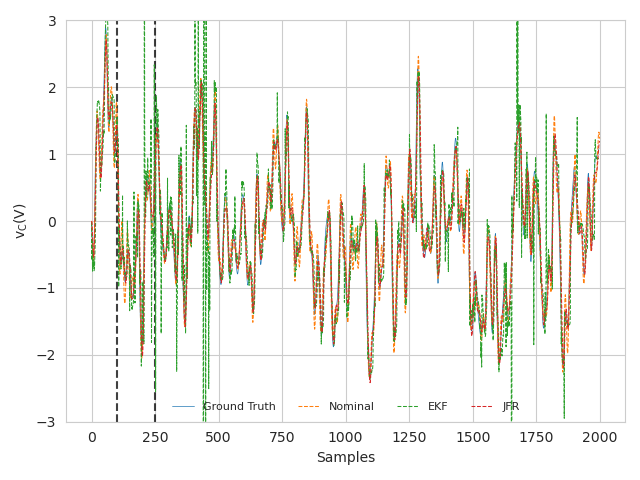}\\
\includegraphics[width=0.70\textwidth]{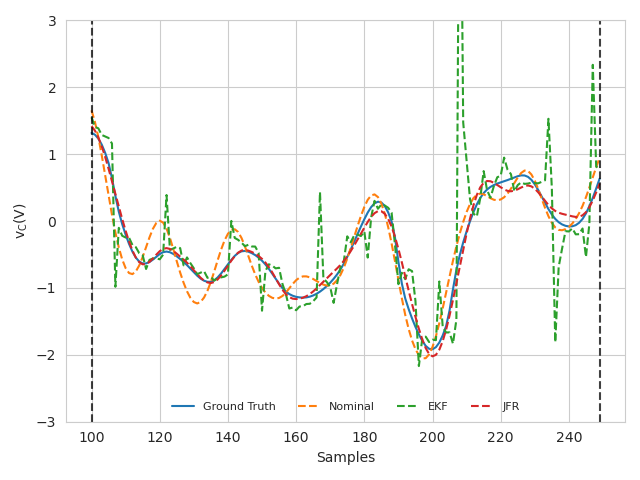}
    \caption{RLC example. \textbf{(Top)} Normalized output voltage $v_C$ on the evaluation dataset: ground truth (blue), nominal state-space neural network~\cite{forgione2021continuous} (orange); Jacobian Feature Regression (red); Extended Kalman filter (green).
    \textbf{(Bottom)} Zoom for samples between the dotted vertical lines (black) in the top figure.}
    \label{RLC-prediction-compare-zoom}
\end{figure}

As shown in Table~\ref{tab:RLC_performance}, the performance of the nominal model is excellent on both
the training and the test datasets,  but drops significantly on the transfer and evaluation datasets where the system is simulated in its perturbed configuration. On the other hand, the JFR model adapted on the transfer dataset achieves very high performance both on the transfer and on the evaluation datasets. Fig.~\ref{RLC-prediction-compare-zoom} shows the (normalized) output voltage $v_C$ on the evaluation dataset for: the nominal state-space network; the JFR model; and an adapted model iteratively estimated through an Extended Kalman Filter (EKF). The obtained results clearly show that the JFR approach proposed in the paper outperforms EKF.

\begin{table}[!tb]
\centering
\caption{RLC example: nominal and perturbed system coefficients.}
\begin{tabular}{c | c  c  c}
\multirow{2}{*}{\textbf{System}} & \multicolumn{3}{c}{\textbf{Parameters}} \\
                & $R$ ($\Omega$)     & $L_0$ ($\mu$H)     & $C$ (nF)     \\ 
\hline
nominal         & $3$  & $50$  & $270$  \\
perturbed       & $4$   & $50$ & $350$ \\
\hline
\end{tabular}
\label{tab:RLC_coefficients}
\end{table}

\begin{table}[!tb]
\centering
\caption{RLC example: $R^2$  performance index  of  nominal and adapted model for training, test, transfer and evaluation datasets.}
\begin{tabular}{l | c  c  c  c}
\multirow{2}{*}{\textbf{Model}} & \multicolumn{4}{c}{\textbf{Dataset}} \\
                & Train      & Test       & Transfer & Eval       \\ 
\hline
nominal         & 0.99       & 0.98       & 0.92     & 0.93 \\
adapted         & -          & -          & 0.99     & 0.97  \\
\hline
\end{tabular}
\label{tab:RLC_performance}
\end{table}

\begin{table*}[h!]
\caption{RLC example. ($i$) $R^2$ performance index on evaluation dataset achieved by: nominal model; Bayesian Linear Regression; Extended Kalman Filter; complete retraining from transfer dataset after full convergence and after $15$ seconds of training;  ($ii$) CPU time T required to compute adapted models through: Bayesian Linear Regression with na{\"i}ve and sensitivity-based (eq.~\eqref{eq:sens}) computation of  the Jacobian; Extended Kalman Filter; full retraining.}
\resizebox{.48\textwidth}{!}{

\begin{tabular}{c | c | c  c | c  c | c  c | c  c  c}
\hline
\hline
\multicolumn{10}{c}{$L_o$ = $60$~$\mu$H , C = $550$~nF \smallskip} \\
\hline
\hline
& \rotatebox[origin=c]{0}{Nominal} &
\multicolumn{2}{c|}{\parbox{1.2cm}{
JFR with \\
na{\"i}ve\\ Jacobian \\ comp. \smallskip}}
& \multicolumn{2}{c|}{\parbox{1.3cm}{\ \\ JFR with \\
recursive\\ Jacobian \\ comp. \smallskip}} & \multicolumn{2}{c|}{EKF} & \multicolumn{3}{c}{Retrain} \\
\hline
& & & & & & & & \multicolumn{2}{c}{full-convergence} & $15s$ \\
\hline
R~($\Omega$) & $R^2$ & $R^2$ & T (s) & $R^2$ & T (s) & $R^2$ & T (s) & $R^2$ & T (s) & $R^2$ \\
\hline
$4$         &    0.80 &    0.95 & 13.52 &   0.95 & 0.98    & 0.68 & 538.80   &  0.98 & 158.02 & 0.07\\
$7$         &    0.73 &    0.94 & 13.14 &   0.94 & 0.97    & 0.83 & 542.21   &  0.99 & 158.32 & 0.20\\
$10$        &    0.66 &    0.91 & 13.06 &   0.91 & 0.99    & 0.85 & 548.64   &  0.99 & 176.42 & 0.15\\
\hline
\end{tabular}
} \hspace*{.04\textwidth}
%
%
\resizebox{.48\textwidth}{!}{
\begin{tabular}{c | c | c  c | c  c | c  c | c c c}
\hline
\hline
\multicolumn{10}{c}{$L_o$ = $60$~$\mu$H , C = $1050$~nF \smallskip} \\
\hline
\hline
 & \rotatebox[origin=c]{0}{Nominal} & %
\multicolumn{2}{c|}{\parbox{1.2cm}{\ \\
JFR with \\
na{\"i}ve\\ Jacobian \\ comp. \smallskip}}
& \multicolumn{2}{c|}{\parbox{1.3cm}{\ \\ JFR with \\
recursive\\ Jacobian \\ comp. \smallskip}} & \multicolumn{2}{c|}{EKF} & \multicolumn{3}{c}{Retrain} \\
\hline
& & & & & & & & \multicolumn{2}{c}{full-convergence} & $15s$ \\
\hline
R~($\Omega$) & $R^2$ & $R^2$ & T (s) & $R^2$ & T (s) & $R^2$ & T (s) & $R^2$ & T(s) & $R^2$ \\
\hline
$4$         &    0.64 &    0.88 & 13.59 &   0.88 & 1.00    & 0.64 & 535.40   &  0.98 & 157.59 & 0.22\\
$7$         &    0.50 &    0.82 & 12.73 &   0.82 & 0.97    & 0.65 & 538.85   &  0.99 & 169.94 & 0.28\\
$10$        &    0.36 &    0.76 & 12.62 &   0.76 & 0.97    & 0.81 & 537.95   &  0.99 & 175.31 & 0.45\\
\hline
\end{tabular}
}

\bigskip

\resizebox{0.48\textwidth}{!}{
\begin{tabular}{c | c | c  c | c  c | c  c | c c c}
\hline
\hline
\multicolumn{10}{c}{$L_o$ = $75$~$\mu$H , C = $550$~nF \smallskip} \\
\hline
\hline
 & \rotatebox[origin=c]{0}{Nominal} & %
\multicolumn{2}{c|}{\parbox{1.2cm}{\ \\
JFR with \\
na{\"i}ve\\ Jacobian \\ comp. \smallskip}}
& \multicolumn{2}{c|}{\parbox{1.3cm}{\ \\ JFR with \\
recursive\\ Jacobian \\ comp. \smallskip}} & \multicolumn{2}{c|}{EKF} & \multicolumn{3}{c}{Retrain} \\
\hline
& & & & & & & & \multicolumn{2}{c}{full-convergence} & $15s$ \\
\hline
R~($\Omega$) & $R^2$ & $R^2$ & T (s) & $R^2$ & T (s) & $R^2$ & T (s) & $R^2$ & T (s) & $R^2$ \\
\hline
$4$         &    0.73 &    0.90 & 13.88 &   0.90 & 1.01    & 0.63 & 550.62   &  0.97 & 144.50 & 0.12\\
$7$         &    0.67 &    0.90 & 13.22 &   0.90 & 1.01    & 0.74 & 544.81   &  0.98 & 140.43 & 0.19\\
$10$        &    0.61 &    0.88 & 12.57 &   0.88 & 0.97    & 0.72 & 537.06   &  0.98 & 159.24 & 0.24\\
\hline
\end{tabular}
} \hspace*{0.04 \textwidth}
%
%
\resizebox{.48\textwidth}{!}{
\begin{tabular}{c | c | c  c | c  c | c  c | c c c}
\hline
\hline
\multicolumn{10}{c}{$L_o$ = $75$~$\mu$H , C = $1050$~nF \smallskip} \\
\hline
\hline
 & \rotatebox[origin=c]{0}{Nominal} & %
\multicolumn{2}{c|}{\parbox{1.2cm}{\ \\
JFR with \\
na{\"i}ve\\ Jacobian \\ comp. \smallskip}}
& \multicolumn{2}{c|}{\parbox{1.3cm}{\ \\ JFR with \\
recursive\\ Jacobian \\ comp. \smallskip}} & \multicolumn{2}{c|}{EKF} & \multicolumn{3}{c}{Retrain} \\
\hline
& & & & & & & & \multicolumn{2}{c}{full-convergence} & $15s$ \\
\hline
R~($\Omega$) & $R^2$ & $R^2$ & T (s) & $R^2$ & T (s) & $R^2$ & T (s) & $R^2$ & T (s) & $R^2$ \\
\hline
$4$         &    0.57 &    0.82 & 13.07 &   0.82 & 0.97    & 0.76 & 540.73   &  0.98 & 390.67 & 0.22\\
$7$         &    0.45 &    0.78 & 12.62 &   0.78 & 0.99    & 0.76 & 537.78   &  0.98 & 148.42 & 0.32\\
$10$        &    0.32 &    0.73 & 12.71 &   0.73 & 0.98    & 0.75 & 543.11   &  0.99 & 166.83 & 0.47\\
\hline
\end{tabular}
}

\bigskip

\resizebox{.48\textwidth}{!}{
\begin{tabular}{c | c | c  c | c  c | c  c | c c c}
\hline
\hline
\multicolumn{10}{c}{$L_o$ = $90$~$\mu$H , C = $550$~nF \smallskip} \\
\hline
\hline
 & \rotatebox[origin=c]{0}{Nominal} & %
\multicolumn{2}{c|}{\parbox{1.2cm}{\ \\
JFR with \\
na{\"i}ve\\ Jacobian \\ comp. \smallskip}}
& \multicolumn{2}{c|}{\parbox{1.3cm}{\ \\ JFR with \\
recursive\\ Jacobian \\ comp. \smallskip}} & \multicolumn{2}{c|}{EKF} & \multicolumn{3}{c}{Retrain} \\
\hline
& & & & & & & & \multicolumn{2}{c}{full-convergence} & $15s$ \\
\hline
R~($\Omega$) & $R^2$ & $R^2$ & T (s) & $R^2$ & T (s) & $R^2$ & T (s) & $R^2$ & T (s) & $R^2$ \\
\hline
$4$         &    0.66 &    0.85 & 12.90 &   0.85 & 0.97    & -82.44 & 537.25   &  0.96 & 390.15 & 0.13\\
$7$         &    0.60 &    0.85 & 13.17 &   0.85 & 0.97    & 0.78 & 540.22   &  0.98 & 133.40 & 0.17\\
$10$        &    0.56 &    0.84 & 12.67 &   0.84 & 0.98    & 0.77 & 538.96   &  0.98 & 147.66 & 0.17\\
\hline
\end{tabular}
} \hspace*{0.04 \textwidth}
%
%
\resizebox{.48\textwidth}{!}{
\begin{tabular}{c | c | c  c | c  c | c  c | c c c}
\hline
\hline
\multicolumn{10}{c}{$L_o$ = $90$~$\mu$H , C = $1050$~nF \smallskip} \\
\hline
\hline
 & \rotatebox[origin=c]{0}{Nominal} & %
\multicolumn{2}{c|}{\parbox{1.2cm}{\ \\
JFR with \\
na{\"i}ve\\ Jacobian \\ comp. \smallskip}}
& \multicolumn{2}{c|}{\parbox{1.3cm}{\ \\ JFR with \\
recursive\\ Jacobian \\ comp. \smallskip}} & \multicolumn{2}{c|}{EKF} & \multicolumn{3}{c}{Retrain} \\
\hline
& & & & & & & & \multicolumn{2}{c}{full-convergence} & $15s$ \\
\hline
R~($\Omega$) & $R^2$ & $R^2$ & T (s) & $R^2$ & T (s) & $R^2$ & T (s) & $R^2$ & T (s) & $R^2$ \\
\hline
$4$         &    0.52 &    0.77 & 13.27 &   0.77 & 0.97    & 0.80 & 539.87   &  0.97 & 391.41 & 0.22\\
$7$         &    0.41 &    0.74 & 13.51 &   0.74 & 0.96    & 0.78 & 536.80   &  0.98 & 133.35 & 0.37\\
$10$        &    0.23 &    0.69 & 13.26 &   0.69 & 0.96    & 0.81 & 538.65   &  0.98 & 153.58 & 0.47\\
\hline
\end{tabular}
}
\label{tab:RLC_sim2real_compare}
\end{table*}


Table \ref{tab:RLC_sim2real_compare} summarizes results of a more extensive validation of the proposed methodology  both in terms of model performance and run time. In particular,  resistance $R$, inductance $L$, and capacitance $C$ are all varied with respect to their nominal values.    
We can observe a drop in the performance of the nominal model on the evaluation dataset with respect to the nominal  performance reported in Table~\ref{tab:RLC_performance} on the test dataset. On the other hand, the proposed  JFR approach for model adaptation achieves consistently better performance, with a  degradation only for large nominal/perturbed system mismatch. It is worth noticing that JFR with recurrent Jacobian computation based on sensitivity equations~\eqref{eq:sens} is about $13$x faster than the na{\"i}ve implementation. 
Furthermore, the performance of JFR are consistently better than the ones achieved by the EKF, both in terms of prediction accuracy and run time. 

Finally,  results achieved by  simply re-training a RNN model from the transfer dataset are also provided.  To appreciate the computational efficiency of the JFR, we compare its performance with  the one achieved by a model retrained for the same amount of time as it takes the JFR with a  na{\"i}ve  computation of the Jacobian  ($\sim$15 seconds). We also report results obtained by retraining the RNN model until full convergence is achieved (either after 10000 iterations or when the loss is smaller than $1$\% of its initial value). Although the predictive performance of the re-trained RNN at full convergence is better than JFR, the run time is about $150$x larger. Furthermore, by constraining the training time to $15$~seconds, we observe a significant drop in the performance of the retrained model. This shows the potential advantage of our JFR approach in presence of time constraints, e.g., in adaptive control applications.

\section{Conclusion}
\label{sec:conclusion}
We have presented a transfer learning methodology for fast and efficient adaptation of Recurrent Neural Network 
models. The identified RNN is augmented by an additive linear-in-the-parameters correction term trained on fresh data using the model's Jacobian features with respect to its nominal parameters. 
The adaptation term is then obtained as the solution of a (non-singular) linear least-squares problem.

The proposed transfer learning approach is shown to compare favorably with respect to EKF in terms of predictive performance and run time. We also compare our methodology with a full   re-identification of the RNN model from scratch. While in some cases full re-identification deliver superior predictive performance, our approach is significantly faster. For a limited computational budget, our approach is actually shown to outperform full plant re-identification also in terms of predictive performance. 
Furthermore, the optimal correction term is guaranteed to exist and to be unique, owing to the linear least-squares formulation.

Current investigations are devoted to the application of our transfer learning methodologies in adaptive control, where the fast run time, the possibility of online implementation, and the theoretical guarantees of the  proposed algorithms may lead to significant advantages with respect to existing model updating techniques.

\section*{Acknowledgments}
The activities of Marco Forgione and Dario Piga have  been supported by HASLER STIFTUNG under the project DEALING: DEep learning for dynamicAL systems and dynamical systems for deep learnING.

\bibliographystyle{unsrt}  
\bibliography{references}

\begin{thebibliography}{10}

\bibitem{schmidhuber2015deep}
J{\"u}rgen Schmidhuber.
\newblock Deep learning in neural networks: An overview.
\newblock {\em Neural networks}, 61:85--117, 2015.

\bibitem{baydin2017automatic}
At{\i}l{\i}m~G{\"u}nes Baydin, Barak~A Pearlmutter, Alexey~Andreyevich Radul,
  and Jeffrey~Mark Siskind.
\newblock Automatic differentiation in machine learning: a survey.
\newblock {\em The Journal of Machine Learning Research}, 18(1):5595--5637,
  2017.

\bibitem{paszke2017automatic}
Adam Paszke, Sam Gross, Soumith Chintala, Gregory Chanan, Edward Yang, Zachary
  DeVito, Zeming Lin, Alban Desmaison, Luca Antiga, and Adam Lerer.
\newblock Automatic differentiation in pytorch.
\newblock 2017.

\bibitem{beintema2021nonlinear}
Gerben Beintema, Roland Toth, and Maarten Schoukens.
\newblock Nonlinear state-space identification using deep encoder networks.
\newblock In {\em Learning for Dynamics and Control}, pages 241--250. PMLR,
  2021.

\bibitem{forgione2021continuous}
M.~Forgione and D.~Piga.
\newblock Continuous-time system identification with neural networks: Model
  structures and fitting criteria.
\newblock {\em European Journal of Control}, 59:69--81, 2021.

\bibitem{andersson2019deep}
Carl Andersson, Ant{\^o}nio~H Ribeiro, Koen Tiels, Niklas Wahlstr{\"o}m, and
  Thomas~B Sch{\"o}n.
\newblock Deep convolutional networks in system identification.
\newblock In {\em 2019 IEEE 58th Conference on Decision and Control (CDC)},
  pages 3670--3676. IEEE, 2019.

\bibitem{forgione2021dynonet}
Marco Forgione and Dario Piga.
\newblock {dynoNet: A} neural network architecture for learning dynamical
  systems.
\newblock {\em International Journal of Adaptive Control and Signal
  Processing}, 35(4), 2021.

\bibitem{iacob2021deep}
Lucian~Cristian Iacob, Gerben~Izaak Beintema, Maarten Schoukens, and Roland
  T{\'o}th.
\newblock {D}eep {I}dentification of {N}onlinear {S}ystems in {K}oopman form.
\newblock {\em arXiv preprint arXiv:2110.02583}, 2021.

\bibitem{mauroy2020koopman}
Alexandre Mauroy, Igor Mezi{\'c}, and Yoshihiko Susuki.
\newblock {\em {T}he {K}oopman {O}perator in {S}ystems and {C}ontrol: Concepts,
  Methodologies, and Applications}, volume 484.
\newblock Springer Nature, 2020.

\bibitem{mavkov2020integrated}
Bojan Mavkov, Marco Forgione, and Dario Piga.
\newblock Integrated neural networks for nonlinear continuous-time system
  identification.
\newblock {\em IEEE Control Systems Letters}, 4(4):851--856, 2020.

\bibitem{pozzoli2020tustin}
Simone Pozzoli, Marco Gallieri, and Riccardo Scattolini.
\newblock Tustin neural networks: a class of recurrent nets for adaptive mpc of
  mechanical systems.
\newblock {\em IFAC-PapersOnLine}, 53(2):5171--5176, 2020.

\bibitem{hastie01statisticallearning}
Trevor Hastie, Robert Tibshirani, and Jerome Friedman.
\newblock {\em The Elements of Statistical Learning}.
\newblock Springer Series in Statistics. Springer New York Inc., New York, NY,
  USA, 2001.

\bibitem{Ljung}
Lennart Ljung.
\newblock {\em System Identification: Theory for the User}.
\newblock Prentice-Hall, Inc., USA, 1986.

\bibitem{maddox2021fast}
Wesley Maddox, Shuai Tang, Pablo Moreno, Andrew~Gordon Wilson, and Andreas
  Damianou.
\newblock Fast adaptation with linearized neural networks.
\newblock In {\em International Conference on Artificial Intelligence and
  Statistics}, pages 2737--2745. PMLR, 2021.

\bibitem{kingma2014adam}
Diederik~P Kingma and Jimmy Ba.
\newblock Adam: A method for stochastic optimization.
\newblock {\em arXiv preprint arXiv:1412.6980}, 2014.

\bibitem{Maciejowski_book}
Jan Maciejowski.
\newblock {\em Predictive Control with Constraints}.
\newblock Prentice Hall, 2000.

\bibitem{Armenio2019}
Luca~Bugliari Armenio, Enrico Terzi, Marcello Farina, and Riccardo Scattolini.
\newblock Echo state networks: analysis, training and predictive control.
\newblock In {\em 2019 18th European Control Conference ({ECC})}. {IEEE}, June
  2019.

\bibitem{rasmussen2006gaussian}
Christopher~K Williams and Carl~Edward Rasmussen.
\newblock {\em Gaussian processes for machine learning}, volume~2.
\newblock MIT press Cambridge, MA, 2006.

\bibitem{balestriero2021fast}
Randall Balestriero and Richard Baraniuk.
\newblock Fast {J}acobian-{V}ector product for {D}eep {N}etworks.
\newblock {\em arXiv preprint arXiv:2104.00219}, 2021.

\bibitem{rabitz1983sensitivity}
Herschel Rabitz, Mark Kramer, and D~Dacol.
\newblock Sensitivity analysis in chemical kinetics.
\newblock {\em Annual review of physical chemistry}, 34(1):419--461, 1983.

\bibitem{nocedal2006numerical}
Jorge Nocedal and Stephen Wright.
\newblock {\em Numerical optimization}.
\newblock Springer Science \& Business Media, 2006.

\bibitem{ljung:1999system}
Lennart Ljung, editor.
\newblock {\em System Identification: Theory for the User}.
\newblock Prentice Hall PTR, Upper Saddle River, NJ, USA, 2 edition, 1999.

\bibitem{CSTR_textbook}
{\em Chemical Reactor Design and Control}, chapter~2, pages 31--106.
\newblock John Wiley \& Sons, Ltd, 2007.

\bibitem{hochreiter1997a}
S.~Hochreiter and J.~Schmidhuber.
\newblock Long short-term memory.
\newblock {\em Neural Computation}, 9(8):1735--1780, 1997.

\bibitem{greff_lstm_2017}
Klaus Greff, Rupesh~Kumar Srivastava, Jan Koutn\'{i}k, Bas~R. Steunebrink, and
  J\"{u}rgen Schmidhuber.
\newblock {LSTM}: {A} {Search} {Space} {Odyssey}.
\newblock {\em IEEE Transactions on Neural Networks and Learning Systems},
  28(10):2222--2232, October 2017.
\newblock arXiv: 1503.04069.

\end{thebibliography}

\end{document}